\pdfoutput=1
\documentclass[11pt]{article}
\usepackage{geometry}
\usepackage{coling2020}
\usepackage{times}
\usepackage{url}
\usepackage{placeins}
\usepackage{latexsym}
\usepackage{microtype}
\usepackage[export]{adjustbox}
\usepackage[title]{appendix}
\usepackage{etoolbox}
\patchcmd{\appendices}{\quad}{. }{}{}
\usepackage{graphicx}
\graphicspath{ {./images/} }

\colingfinalcopy 

\title{Yseop at SemEval-2020 Task 5: Cascaded BERT Language Model for Counterfactual Statement Analysis}

\author{Hanna Abi Akl \\
  Yseop \\
  {\tt habi-akl@yseop.com} \\\And
  Dominique Mariko \\
  Yseop \\
  {\tt dmariko@yseop.com} \\\And
  Estelle Labidurie \\
  Yseop \\
  {\tt elabidurie@yseop.com}}

\date{}

\begin{document}
\maketitle
\begin{abstract}
In this paper, we explore strategies to detect and evaluate counterfactual sentences. We describe our system for SemEval-2020 Task 5: Modeling Causal Reasoning in Language: Detecting Counterfactuals. We use a BERT base model for the classification task and build a hybrid BERT Multi-Layer Perceptron system to handle the sequence identification task. Our experiments show that while introducing syntactic and semantic features does little in improving the system in the classification task, using these types of features as cascaded linear inputs to fine-tune the sequence-delimiting ability of the  model ensures it outperforms other similar-purpose complex systems like BiLSTM-CRF in the second task. Our system achieves an F1 score of {85.00\%} in Task 1 and {83.90\%} in Task 2. 

\end{abstract}

\section{Introduction}
\label{intro}

%
%
\blfootnote{
    %
    %
    %
    %
     \hspace{-0.65cm}  
     This work is licensed under a Creative Commons 
     Attribution 4.0 International Licence.
     Licence details:
     \url{http://creativecommons.org/licenses/by/4.0/}.
    %
    %
}

A counterfactual can be defined as something that is contrary to the truth or that did not actually occur. It refers to an event that did not or cannot happen, as well as the possible consequences if it had happened. In the sentence \textit{"If dogs had no ears, they could not hear"}, the statement \textit{"if dogs had no ears"} is an example of a counterfactual because dogs \textit{do} have ears. Task 5 of SemEval-2020 \cite{Yang:20} focuses on identifying these specific sentence types amongst sentences delivering close semantic similarities. This implies understanding and disambiguating the causal link between two sentence fragments. 

We approached this task as an opportunity to test the effectiveness of disambiguating at a grammatical level against traditional baseline systems. This paper describes a parallel approach in deriving some meaning from text to leverage the influence of context and relevance of structure in recognizing counterfactual statements. Specifically, we will explore how many expressions of such statements a mapping of grammatical types can cover before falling short to high-performing models, most notably BERT \cite{Devlin:18}, which performed well on both tasks with an F1 score of 85.00\% on Task 1 and 83.90 \% on Task 2.

\section{Related Work}
Although the task of detecting counterfactuals is relatively new, \cite{Son:17} proposes using modal logic to form rule-based determination methods from social media posts. These methods are supplemented by a statistical classifier (Linear SVM) that is retrained to tackle more challenging counterfactual forms.

Previous work on causal identification by \cite{Levin:15} studied the contribution of verbs in the determination of causal relations. By analyzing closely different formulations, they concluded that similarities in meaning can be derived from very different syntactic structures

As for causality relation extraction, different deep learning systems built on the success of neural-based models have been proposed. The linguistically informed CNN model \cite{Dasgupta:18} leverages the use of word embeddings and other linguistic features to detect causal patterns and outperforms rule-based classification. \cite{Liang:19} introduces a multi-level causal detector that makes use of multi-head self-attention to capture semantic features at word level and infer causality at segment level. This engineered system has rivaled state-of-the-art models in terms of performance and thorough understanding of complex semantic information such as discourse relations and transitivity rules. Finally, \cite{Li:19} presents a self-attentive BiLSTM-CRF model that makes use of transfer learning to overcome the problem of data insufficiency and extract causal relations in natural language text. This solution transfers a trained embedding from a large corpus and uses the causality tagging scheme to identify dependencies between cause and effect. Experimental results prove the effectiveness of this model, but its major limitation is the insufficiency of high-quality annotated data to learn from.     

\section{Dataset}
The datasets used are those provided by the shared task organisers. The data is described in \cite{Yang:20}. As per official task instructions, no additional data was used.

\section{Task 1: Classification Problem}
Task 1 is a classification problem which aims at recognising text sections as either \textit{counterfactual} or \textit{not}. Counterfactual sections are labeled \textbf{1} and non-counterfactual sections are labeled \textbf{0}.

The proposed baseline by the task organisers is a Support Vector Machine (SVM) classifier. After a preliminary phase based on exploring grammatical features engineering, we discuss two different approaches and compare them to the baseline: a classical approach using popular machine learning classifiers with semantic and additional grammatical features, a combination that has shown to perform well on information retrieval tasks relying on text understanding \cite{Dai:19}, and a deep learning approach using a BERT linear classifier. The objective of this competitive approach is to determine whether a supplement of linguistic features can be sufficient to correctly recognise counterfactual structures as opposed to running heavier models integrating broader contextual knowledge.

\subsection{Proposed Method}
\label{sect:method1}
\subsubsection{Exploration}
As the data set contains sentences with close semantics, we run a first exploratory analysis to try to establish disambiguation patterns. A manual linguistic analysis of the training dataset shows that verbs are key elements for the detection of counterfactuals. \cite{Son:17} identify 7 characteristics related to counterfactuals, all depending on a verb feature. Verb tenses in particular are key to disambiguation, so we build a generalist grammar based on our observation to include only the verbal forms that seem relevant for our analysis. We categorize \textit{could}, \textit{would} and \textit{should} as modals so they wouldn't be identified as verbs in the preterit tense. Finally, we add a pattern to disambiguate the category of \textit{wish (verb)} and \textit{wishes (noun)}, and identify conditional statements \textit{(If)}.
We eventually retain 4 main features among the 7 described by \cite{Son:17}: Verb inversion, Modal chunks, Wish verbs and If clauses. We then generate combinations of tokens based on their grammatical category (i.e., verb, pronoun) as well as their linguistic properties (i.e., tense). The full grammar is provided in Appendix A.

However, some cases resist disambiguation, especially the sentences containing \textit{could}/\textit{would}/\textit{should have} structures, which are very common in English and in many cases do not imply a counterfactual. Moreover, the context of these structures can vary which makes them even more complex to disambiguate. 
Eventually, the ambiguity between adjectives and past participles can raise many mistakes in the identification of counterfactuals when using grammars. Adjectives constructed with the suffix \textit{-ed} are also recognised as verbs in preterit tense or past participles by the grammar (i.e., \textit{He wasn't prepared}).

By deterministically applying the grammar on the training set, we obtain an output of 2833 sentences recognised as counterfactuals (on a total of 13000). However, only 593 of them are labeled "1" in the training set. A manual analysis of the output reveals that many statements are recognised as counterfactuals because of remaining ambiguities.

We turn to classical linear and non-linear machine learning methods to explore if combining these patterns carries any prior knowledge on disambiguation. We turn all 4 features described in Appendix A into binary variables, flagging the existence (True) or non-existence (False) of the category and associated patterns in the sentence. An example of the data transformation is provided in Appendix B. We end up with a table of 31 features that are evaluated for predicting a counterfactual statement with the following learning methods and parameters (using scikit-learn classifiers\footnote{https://github.com/scikit-learn/scikit-learn}):

\par \textbf{SVM:} \textit{gamma}: scale \
\par \textbf{LOGIT:} \textit{l1 ratio}: 0.5, \textit{penalty}: elasticnet, \textit{solver}: saga \
\par \textbf{KNN:} \textit{n neighbors}: 3 \
\par \textbf{CART:} \textit{criterion}: gini \
\par \textit{all other parameters:} sklearn defaults \\
\begin{table}[h]
\begin{center}
\begin{tabular}{|l|r|r|r|r|r|r|}
\hline \bf Classifier &  \bf True value &  \bf Precision & \bf Recall & \bf F1 score \\ \hline
SVM & 0 & 0.89 & 1.00 & 0.94  \\
& 1 & 0.70 & 0.05 & 0.10  \\
\hline
LOGIT & 0 & 0.89 & 1.00 & 0.94  \\
& 1 & 0.71 & 0.06 & 0.11 \\
\hline
KNN & 0 & 0.90 & 0.98 & 0.94 \\
& 1 & 0.39 & 0.12 & 0.18 \\
\hline
CART & 0 &  0.89  & 1.00 & 0.94 \\
& 1 &  0.68 & 0.06 & 0.11 \\

\hline
\end{tabular}
\end{center}
\caption{\label{font-table} Task 1 - Exploratory Tests}
\end{table}

The results provided in Table 1, especially the very poor recall metrics, prove our initial set of features are incomplete and noisy.
We then decide for a more holistic approach, resuming any prior knowledge on the syntax to evaluate if existing latent variables can possibly lie in the raw morpho-syntax (POS tags) and semantic of counterfactual statements.

\subsubsection{Operating methods}
We first try a classical approach, supplementing classifiers with word vectorisation (Bag Of Words (BOW), TF-IDF, Word2Vec and BERT vectors) and morpho-syntactic features (POS tagging). The details of these transformations are provided in Appendix C. The classifiers evaluated are: SVM, K-Nearest-Neighbor (KNN), Multinomial Naive Bayes (NB), Decision Trees (CART), Random Forest (RF) and Multi-Layer Perceptron (MLP). Finally, we add a cross validation method with a stratified fold of 3 to our models.
As shown in Table 2, the best learning models seem to have strong linearisation capabilities (MLP and SVM), which is why we also tested a deep learning model which has proved similar behaviour.

For the deep learning approach, we use a model inspired from the BERT model that achieved the best result in a similar text classification task in the NLP4IF-2019 Shared Task \cite{Da:19}. This is a BERT model with a linear layer on top.
This system has proven to pay special attention to adjectives and verbs, two grammatical categories that can play a role in identifying counterfactual statements \cite{Levin:15}.
For our implementation, we use the \textit{BertForSequenceClassification} model from the Transformers \footnote{https://github.com/huggingface/transformers} library, a sentence-tokenized version of the BERT Uncased model with 12 Transformer layers and 110 million parameters. 

\subsection{Results}
\label{sect:results1}
The results are based on the official training set provided by the organisers. The dataset contains 13000 lines split in the following: 40\% training sample, 30\% validation sample and 30\% test sample.

For the classical approach, we retain the best result for each classifier out of all the possible combinations of classifiers and text processing and compare with the baseline provided for the task and the deep learning model. The +/- scores are the averaged measures from the 3-fold cross-validation results for all models except the BERT Linear Model, whose results are provided with the scikit-learn default parameters.

\begin{table}[h]
\begin{center}
\begin{tabular}{|l|r|r|r|r|r|}
\hline \bf Classifier &  \bf Processing &  \bf Precision & \bf Recall & \bf F1 score \\ \hline
Baseline SVM & TF-IDF & 72.72 \% & 8.73 \% & 15.59 \% \\
MLP & BERT Sentence Version & 80 +/- 1 \% & 80 +/- 1 \% & 80 +/- 1 \% \\
SVM & BERT Sentence Version & 79 +/- 0 \% & 81 +/- 1 \% & 80 +/- 0 \% \\
CART & TF-IDF with Stop Words & 69 +/- 1 \% & 70 +/- 1 \% & 70 +/- 1 \% \\
NB & BERT Sentence Version & 64 +/- 1 \% & 75 +/- 2 \% & 66 +/- 1 \% \\
KNN & TF-IDF with Stop Words & 71 +/- 1 \% & 64 +/- 1 \% & 66 +/- 1 \% \\
RF & TF-IDF with Stop Words & \textbf{91 +/- 1 \%} & 61 +/- 1 \% & 65 +/- 1 \% \\
BERT Linear & BERT Uncased & 87 +/- 1 \% & \textbf{83 +/- 1 \%} & \textbf{86 +/- 1 \%} \\
\hline
\end{tabular}
\end{center}
\caption{\label{font-table} Task 1 - Benchmark Tests}
\end{table}

The blind test set consists of 7000 unlabeled lines of text. Our best model, the BERT Linear, achieves an F1 score of \textbf{85.00\%}, a Precision score of \textbf{84.20\%} and a Recall score of \textbf{85.90\%} on this set.

\section{Task 2: Sequence Delimitation}
\label{ssec:task2}
Task 2 is a sequence delimitation problem. The text sections are similar to the ones labeled "1" in the Task 1 dataset.
The purpose of this task is to extract, in a text section identified as counterfactual, the sub-strings identifying the antecedent and the consequent elements \cite{Yang:20}. 
We use the following split sampling for the training data (3551 individuals): \textbf{1740 sentences} for the training sample, \textbf{746 sentences} for the validation sample and \textbf{1065 sentences} for the test sample.

\subsection{Proposed Method}

Our approach for this task is also comparative and evaluates two deep learning models. It consists in testing whether we can supply enough linguistic knowledge to determine all causal forms in counterfactual statements to challenge the breadth and depth of a model that leverages the power of BERT.

For both systems, we designed each provided statement in the training set as chunks. These chunks are composed of tokens labelled as  \textit{C} when the token belongs to a sub-string \textit{Consequent} and \textit{A} when the token belongs to a sub-string \textit{Antecedent}. Tokens belonging to neither are marked \textit{I}. 
These transformations of the target sequences, additional transformations and their input levels are described in Appendix D  (D.0.1 and D.0.2). These target features are identified as \textbf{CHUNKS} in our results Table 3 and 4.
Since we can tackle this task through token classification, we build a first Sequence Extractor system using the BERT model from task 1 and a wrapper\footnote{https://github.com/charles9n/bert-sklearn} to concatenate a Multi-Layer Perceptron classifier layer on top of it, as demonstrated in \cite{Dai:19}.

The second system is inspired by discriminative models and Conditional Random Fields (CRF) in particular. We design the system as a Named Entity tagger that takes sentence tokens and a set of morpho-syntactic token-level features as input and predicts the target class of each token. 

We enhance the discriminative properties of the CRF by working on some additional layers and modeling a Deep Learning CRF. We experiment with linguistic embeddings, features and regularisation methods as enhancements for the final BiLSTM-CRF Neural Model. A high-level diagram of the system architecture is presented in Appendix E.

The full configurations for both systems are detailed in Appendix D (D.0.3 and D.0.4).

\subsection{Results}
Since we are working with a complex model with high tuning capabilities, the most relevant results for the BiLSTM-CRF system on the training dataset are retained and presented in Table 4. The BERT-MLP model results are shown in Table 5. Here again the +/- measures are averaged from the 3-fold cross validation scores.
\begin{table}[h]
\begin{center}
\adjustbox{max width=\textwidth}{%
\begin{tabular}{|l|r|r|r|r|r|r|r|}
\hline \bf Model & \bf Embedding & \bf Features & \bf Regularisation &  \bf Precision & \bf Recall & \bf F1 score \\ \hline
BiLSTM-CRF & FastText + C2IDX & POS + CHUNKS & B & 77 +/- 1  \% & 76 +/- 2 \% & 77 +/- 2 \% \\
BiLSTM-CRF & Stacked + C2IDX & POS + CHUNKS + SG & C & 76 +/- 1 \% & 75 +/- 1 \% & 77 +/- 2 \% \\
BiLSTM-CRF & GloVe + C2IDX & POS + CHUNKS & C & \textbf{78 +/- 1 \%} & \textbf{78 +/- 1 \%} & \textbf{79 +/- 1 \%} \\
BiLSTM-CRF & GloVe + C2IDX & POS + CHUNKS + SG & A & 76 +/- 1 \% & 75 +/- 1 \% & 77 +/- 1 \% \\
BiLSTM-CRF & C2IDX & POS + CHUNKS + BERTvec + SG & B & 76 +/- 1 \% & 76 +/- 1 \% & 77 +/- 1 \% \\
BiLSTM-CRF & Stacked + C2IDX & POS + CHUNKS & C & 76 +/- 1 \% & 76 +/- 1 \% & 76 +/- 1 \% \\
\hline
\end{tabular}}
\end{center}
\caption{\label{font-table} Task 2 - BiLSTM-CRF Benchmark Tests}
\end{table}

\begin{table}[h]
\begin{center}
\adjustbox{max width=\textwidth}{%
\begin{tabular}{|l|r|r|r|r|r|r|}
\hline \bf Model & \bf Features & \bf Regularisation & \bf Precision & \bf Recall & \bf F1 score \\ \hline
BERT-MLP & POS + CHUNKS & A & 81 +/- 1  \% & 82 +/- 1 \% & 81 +/- 1 \% \\
BERT-MLP & CHUNKS + SG & A & 82 +/- 1 \% & 83 +/- 1 \% & 82 +/- 1 \% \\
BERT-MLP & CHUNKS & A & \textbf{85 +/- 1 \%} & \textbf{85 +/- 1 \%} & \textbf{85 +/- 1 \%} \\
BERT-MLP & CHUNKS + BIO NER & A & 83 +/- 1 \% & 84 +/- 1 \% & 83 +/- 1 \% \\
BERT-MLP & CHUNKS + SG + BIO NER & A & 83 +/- 1 \% & 83 +/- 1 \% & 83 +/- 1 \% \\
BERT-MLP & POS + CHUNKS + SG + BIO NER & A & 84 +/- 1 \% & 84 +/- 1 \% & 84 +/- 1 \% \\
\hline
\end{tabular}}
\end{center}
\caption{\label{font-table} Task 2 - BERT-MLP Benchmark Tests}
\end{table}

The results show that generalization might be a problem for both systems as discussed in \cite{Zey:19}. The gap in performance between the BiLSTM-CRF and BERT-MLP is due to the superior performance of Transformer models over LSTM encoders. When coupling the BERT-MLP model with one set of features, we observe steep learning, but cascading multiple feature layers results in a smoother and more accurate learning process. Ultimately, since the accuracy of the other linguistic features is closely tied to the library that generates them (particularly the POS tagger), they do not add any enhancement to the chunking feature. 
On the blind test dataset, our best model achieves a Precision score of \textbf{82.30\%}, a Recall score of \textbf{88.80\%}, an F1 score of \textbf{83.90\%} and an Exact Match of \textbf{25.90\%}. The Exact Match measure is severely affected by chunking limits which explains why the score for this metric is particularly poor.

\section{Conclusion}
In this paper, we presented our experiments for identifying counterfactual statements. Modeling features derived from a linguistic analysis such as specific grammar structures for counterfactual statements and coupling them with established machine learning or deep learning models did not perform as well as context-learning models, as our hybrid BERT-MLP solution outperforms even complex combinations of deep learners and displays a better level of understanding and handling challenging counterfactual forms. Future work could explore the impact of graph knowledge in accommodating systems and rendering them more perceptive of implicit and ambiguous textual meanings.

\begin{appendices}

\section{List of disambiguation features and associated grammars.}
We qualify the 4 identified features using structures which are sequences of Part-Of-Speech tags. The Part-of-Speech tags follow the Penn Treebank convention and are extracted using Stanford CoreNLP\footnote{https://stanfordnlp.github.io/CoreNLP/index.html}.
Features are listed along with their corresponding sequences and example sentences. The grammar is implemented with NLTK.

\FloatBarrier
\begin{table}[h!]
\begin{center}
\adjustbox{max width=\textwidth}{%
\begin{tabular}{|l|r|r|}
\hline \bf Feature & Pattern & \bf Example \\ \hline

\textbf{Verb inversion} & {$<$VBD$>$ $<$PRP$>$ $<$VBN$>$} & (\textit{Had\underline{/VBD} you\underline{/PRP} provided\underline{/VBN}}) \\

& {$<$VBD$>$ $<$PRP$>$ $<$RP$>$ $<$VBN$>$} & (\textit{Had\underline{/VBD} you\underline{/PRP} not\underline{/RP} provided\underline{/VBN}}) \\

& {$<$VBD$>$ $<$PRP$>$ $<$VBD$>$} & (\textit{Had\underline{/VBD} she\underline{/PRP} asked\underline{/VBD}}) \\

& {$<$VBD$>$ $<$PRP$>$ $<$RP$>$ $<$VBD$>$} & (\textit{Had\underline{/VBD} she\underline{/PRP} not\underline{/RP} asked\underline{/VBD}}) \\

\hline
\textbf{Modal chunks} & {$<$MD$>$ $<$VB$>$ $<$VBN$>$} & (\textit{Would\underline{/MD} have\underline{/VB} provided\underline{/VBN}}) \\

& {$<$MD$>$ $<$VB$>$ $<$VB$>$} & (\textit{Would\underline{/MD} go\underline{/VB} check\underline{/VB}}) \\

& {$<$MD$>$ $<$RP$>$ $<$VB$>$ $<$VBD$>$} & (\textit{Would\underline{/MD} not\underline{/RP} have\underline{/VB} died\underline{/VBD}}) \\

& {$<$MD$>$ $<$VB$>$} & (\textit{Would\underline{/MD} be\underline{/VB}}) \\

& {$<$MD$>$ $<$RP$>$ $<$VB$>$} & (\textit{Should\underline{/MD} not\underline{/RP} have\underline{/VB}}) \\

& {$<$MD$>$ $<$VB$>$ $<$VBD$>$} & (\textit{Would\underline{/MD} have\underline{/VB} asked\underline{/VBD}}) \\

& {$<$MD$>$ $<$VBP.*$>$ $<$VBN$>$} & (\textit{Would\underline{/MD} have\underline{/VBP} bought\underline{/VBN}}) \\

& {$<$MD$>$ $<$RP$>$ $<$VBP.*$>$ $<$VBN$>$} & (\textit{Would\underline{/MD} not\underline{/RP} have\underline{/VBP} bought\underline{/VBN}}) \\

& {$<$MD$>$ $<$RP$>$ $<$VB$>$ $<$VBN$>$} & (\textit{Would\underline{/MD} not\underline{/RP} have\underline{/VB} provided\underline{/VBN}}) \\

& {$<$MD$>$ $<$VBP.*$>$ $<$VBN$>$ $<$VBN$>$} & (\textit{Would\underline{/MD} have\underline{/VBP} been\underline{/VBN} bought\underline{/VBN}}) \\

& {$<$MD$>$ $<$RP$>$ $<$VBP.*$>$ $<$VBN$>$ $<$VBN$>$} & (\textit{Would\underline{/MD} not\underline{/RP} have\underline{/VBP} been\underline{/VBN} bought\underline{/VBN}}) \\
\hline
\textbf{Wish verbs} & {$<$PRP$>$ $<$VBP$>$ $<$PRP.*$>$ $<$VBD$>$} & (\textit{I\underline{/PRP} wish\underline{/VBP} I\underline{/PRP} held\underline{/VBD}}) \\

& {$<$PRP$>$ $<$VBP$>$ $<$IN$>$ $<$PRP$>$ $<$VBD$>$} & (\textit{I\underline{/PRP} wish\underline{/VBP} that\underline{/IN} you\underline{/PRP} had\underline{/VBD}}) \\

& {$<$PRP$>$ $<$VBP$>$ $<$IN$>$ $<$PRP$>$ $<$VBD$>$ $<$VBN$>$} & (\textit{I\underline{/PRP} wish\underline{/VBP} that\underline{/IN} I\underline{/PRP} had\underline{/VBD} bought\underline{/VBN}}) \\

& {$<$MD$>$ $<$VB$>$ $<$IN$>$ $<$PRP$>$ $<$VBD$>$ $<$VBN$>$} & (\textit{May\underline{/MD} wish\underline{/VB} that\underline{/IN} he\underline{/PRP} had\underline{/VBD} kept\underline{/VBN}}) \\

& {$<$MD$>$ $<$VB$>$ $<$IN$>$ $<$NN.*$>$ $<$VBD$>$ $<$VBN$>$} & (\textit{May\underline{/MD} wish\underline{/VB} that\underline{/IN} households\underline{/NNS} had\underline{/VBD} kept\underline{/VBN}}) \\

& {$<$MD$>$ $<$VB$>$ $<$IN$>$ $<$DT$>$ $<$NN.*$>$ $<$VBD$>$ $<$VBN$>$} & (\textit{May\underline{/MD} wish\underline{/VB} that\underline{/IN} these\underline{/DT} households\underline{/NNS} had\underline{/VBD} kept\underline{/VBN})} \\

& {$<$PRP$>$ $<$VBP$>$ $<$PRP$>$ $<$VBD$>$} & (\textit{I\underline{/PRP} wish\underline{/VBP} you\underline{/PRP} had\underline{/VBD})} \\

& {$<$PRP$>$ $<$DT$>$ $<$VBP$>$ $<$IN$>$ $<$PRP$>$ $<$MD$>$ $<$VB$>$ $<$VBN$>$} & (\textit{We\underline{/PRP} both\underline{/DT} wish\underline{/VBP} that\underline{/IN} we\underline{/PRP} could\underline{/MD} have\underline{/VB} helped\underline{/VBN}}) \\                                           
& {$<$PRP$>$ $<$DT$>$ $<$VBP$>$ $<$PRP$>$ $<$MD$>$ $<$VB$>$ $<$VBN$>$} & (\textit{We\underline{/PRP} both\underline{/DT} wish\underline{/VBP} we\underline{/PRP} could\underline{/MD} have\underline{/VB} helped\underline{/VBN})} \\

& {$<$PRP$>$ $<$VBP$>$ $<$DT$>$ $<$VBD$>$} & (\textit{I\underline{/PRP} wish\underline{/VBP} this\underline{/DT} were\underline{/VBD})} \\

& {$<$PRP$>$ $<$RB$>$ $<$VBP$>$ $<$PRP$>$ $<$MD$>$ $<$VB$>$} & (\textit{I\underline{/PRP} just\underline{/RB} wish\underline{/VBP} we\underline{/PRP} could\underline{/MD} have\underline{/VB}}) \\

& {$<$PRP$>$ $<$RB$>$ $<$VBP$>$ $<$PRP$>$ $<$VBD$>$} & (\textit{I\underline{/PRP} just\underline{/RB} wish\underline{/VBP} we\underline{/PRP} had\underline{/VBD})} \\

& {$<$PRP$>$ $<$RB$>$ $<$VBP$>$ $<$PRP$>$ $<$MD$>$ $<$VB$>$} & (\textit{I\underline{/PRP} just\underline{/RB} wish\underline{/VBP} we\underline{/PRP} could\underline{/MD} have\underline{/VB}}) \\

& {$<$PRP$>$ $<$VBP$>$ $<$PRP.*$>$ $<$MD$>$ $<$VB$>$} & (\textit{I\underline{/PRP} wish\underline{/VBP} we\underline{/PRP} could\underline{/MD} have\underline{/VB})} \\

& {$<$PRP$>$ $<$VBP$>$ $<$PRP.*$>$ $<$NN.*$>$ $<$MD$>$ $<$VB$>$} & (\textit{I\underline{/PRP} wish\underline{/VBP} my\underline{/PRP\$} parents\underline{/NNS} could\underline{/MD} have\underline{/VB}}) \\

& {$<$PRP$>$ $<$VBP$>$ $<$NN.*$>$ $<$MD$>$ $<$VB$>$} & (\textit{I\underline{/PRP} wish\underline{/VBP} institutions\underline{/NNS} could\underline{/MD} have\underline{/VB})} \\

& {$<$PRP$>$ $<$VBP$>$ $<$PRP.*$>$ $<$NN.*$>$ $<$VBD$>$ $<$VBN$>$} & (\textit{I\underline{/PRP} wish\underline{/VBP} my\underline{/PRP\$} parents\underline{/NNS} had\underline{/VBD} known\underline{/VBN}}) \\

& {$<$PRP$>$ $<$VBP$>$ $<$PRP.*$>$ $<$VBD$>$ $<$VBN$>$} & (\textit{I\underline{/PRP} wish\underline{/VBP} they\underline{/PRP} had\underline{/VBD} known\underline{/VBN})} \\

& {$<$PRP$>$ $<$VBP$>$ $<$NN.*$>$ $<$VBD$>$} & (\textit{I\underline{/PRP} wish\underline{/VBP} articles\underline{/NNS} addressed\underline{/VBD}}) \\

& {$<$PRP$>$ $<$VBP$>$ $<$DT$>$ $<$NN.*$>$ $<$VBD$>$} & (\textit{I\underline{/PRP} wish\underline{/VBP} this\underline{/DT} article\underline{/NN} addressed\underline{/VBD}}) \\

& {$<$PRP$>$ $<$VBP$>$ $<$NN.*$>$ $<$VBD$>$ $<$VBN$>$} & (\textit{I\underline{/PRP} wish\underline{/VBP} President\underline{/NNP} Obama\underline{/NNP} had\underline{/VBD} succeeded\underline{/VBN}}) \\

& {$<$PRP$>$ $<$VBP$>$ $<$NN.*$>$ $<$VBD$>$} & (\textit{I\underline{/PRP} wish\underline{/VBP} President\underline{/NNP} Obama\underline{/NNP} had\underline{/VBD}}) \\

& {$<$PRP$>$ $<$VBP$>$ $<$PRP$>$ $<$VBD$>$ $<$RP$>$} & (\textit{You\underline{/PRP} wish\underline{/VBP} you\underline{/PRP} had\underline{/VBD} n't\underline{/RP}}) \\

& {$<$PRP$>$ $<$VBP$>$ $<$DT$>$ $<$VBD$>$ $<$JJ$>$} & (\textit{I\underline{/PRP} wish\underline{/VBP} this\underline{/DT} were\underline{/VBD} true\underline{/JJ}}) \\

\hline
\textbf{If clauses} & {$<$IN$>$ $<$PRP$>$ $<$MD$>$ $<$VB$>$} & (\textit{If\underline{/IN} I\underline{/PRP} could\underline{/MD} go\underline{/VB}}) \\

& {$<$IN$>$ $<$PRP$>$ $<$VBD$>$ $<$JJ$>$} & (\textit{If\underline{/IN} it\underline{/PRP} were\underline{/VBD} true\underline{/JJ}}) \\

& {$<$IN$>$ $<$EX$>$ $<$VBD$>$} & (\textit{If\underline{/IN} there\underline{/EX} were\underline{/VBD}}) \\

& {$<$IN$>$ $<$NN.*$>$ $<$VBD$>$} & (\textit{If\underline{/IN} money\underline{/NN} was\underline{/VBD})} \\

& {$<$VBP$>$ $<$PRP$>$ $<$VBN$>$} & (\textit{Have\underline{/VBP} you\underline{/PRP} been\underline{/VBN}}) \\

& {$<$IN$>$ $<$RP$>$ $<$IN$>$ $<$DT$>$} & (\textit{If\underline{/IN} not\underline{/RP} for\underline{/IN} that\underline{/DT}}) \\

& {$<$IN$>$ $<$DT$>$ $<$NN.*$>$ $<$MD$>$ $<$VB$>$} & (\textit{If\underline{/IN} the\underline{/DT} boy\underline{/NN} could\underline{/MD} go\underline{/VB}}) \\

& {$<$IN$>$ $<$DT$>$ $<$JJ.*$>$ $<$NN.*$>$ $<$MD$>$ $<$VB$>$} & (\textit{If\underline{/IN} the\underline{/DT} young\underline{/JJ} boy\underline{/NN} could\underline{/MD} go\underline{/VB}}) \\

& {$<$IN$>$ $<$DT$>$ $<$NN.*$>$ $<$VBD$>$} & (\textit{If\underline{/IN} the\underline{/DT} boy\underline{/NN} was\underline{/VBD})} \\

& {$<$IN$>$ $<$DT$>$ $<$NN.*$>$ $<$VBD$>$} & (\textit{If\underline{/IN} his\underline{/DT} money\underline{/NN} was\underline{/VBD}}) \\ 
\hline
\end{tabular}}
\end{center}
\caption{\label{font-table} Task 1 - Exhaustive feature grammar}
\end{table}
\FloatBarrier

\section{Feature engineering from Appendix A grammar.}
We transform the different representations of the 4 grammatical features listed in Appendix A into categorical variables. Table 6 illustrates an example of one feature representation applied to input sentences.

\FloatBarrier
\begin{table}[h!]
\begin{center}
\adjustbox{max width=\textwidth}{%
\begin{tabular}{|l|r|r|r|r|r|r|r|r|r|r|}
\hline
sentenceID & gold\_label & IF \newline $_{I_{N} D_{T} N_{N} V_{BD}}$ & IF \newline $_{I_{N} D_{T} N_{NP} V_{BD}}$ & IF \newline $_{I_{N} D_{T} N_{NPS} V_{BD}}$ &  IF \newline $_{I_{N} D_{T} N_{NS} V_{BD}}$ & IF \newline $_{I_{N} E_{X} V_{BD}}$ & ...\\ \hline
100000 & 1 & 0 & 0 & 0 & 1 & 0 & ... \\ \hline
100001 & 1 & 0 & 0 & 0 & 0 & 1 & ... \\ \hline
\end{tabular}}
\end{center}
\caption{\label{font-table} Task 1 - Feature engineering examples for If clauses }
\end{table}
\FloatBarrier

\section{Data transformations for Task 1.}

We perform a two-level processing on the input: vectorial and morpho-syntactic. The vectorisation methods used are: Bag Of Words (BOW), TF-IDF, Word2Vec and BERT vectors. We perform a series of operations on the raw input text like removing numbers, punctuation and stop words, replacing negative contraction verbs with their complete forms (i.e., \textit{won't}), splitting compound forms (i.e., \textit{state-of-the-art}) and transforming text to lowercase. We also replace multiple white spaces with a single space. Examples of these transformations are provided in Table 7. For TF-IDF and Word2Vec, we experiment with and without stop word removal. We use the NLTK\footnote{https://github.com/nltk/nltk} stop words set and increase it with contraction patterns like \textit{'re} or \textit{'m}. For the BERT and Word2Vec vectors, we refrain from applying these cleaning operations to maintain more semantic freedom.
 
At the end of this phase, we generate a list of all the unique words in the training data called the vocabulary.  

For the morpho-syntactic phase, we apply POS-based stemming and lemmatisation for the BOW and TF-IDF embeddings. We also remove words with frequency less than 5 for these embeddings. This effectively decreases the dimensions of BOW and TF-IDF vectors.

Our final feature list consists of 2000 features that are the unique lemmatized vocabulary words and word groups curated from the input text. However, despite their simplicity and low time complexity, BOW and TF-IDF have two major drawbacks. First, as the size of the data and the number of unique words in the training text increase, the length of vectors becomes much larger. Moreover, in these two approaches, only words and their repetitions are important and the order of the words in the text is not considered in the model. This is why we also consider both Word2Vec and BERT embedding approaches in our experiments.

\FloatBarrier
\begin{table}[h!]
\begin{center}
\adjustbox{max width=\textwidth}{%
\begin{tabular}{|l|r|r|}
\hline
Transformation & Example \\ \hline
No transformation & If the lawsuit can be another means of focusing attention on these fundamental issues, then optimistically the lawsuit can provide a larger benefit. \\ \hline
Cleaning & if the lawsuit can be another means of focusing attention on these fundamental issues then optimistically the lawsuit can provide larger benefit \\ \hline
Cleaning + Normalization & If the lawsuit can be anoth mean of focu attent on these fundament issu , then optimist the lawsuit can provid a larg benefit . \\ \hline
Cleaning (Word2Vec + No Stop Words) & lawsuit another means focusing attention fundamental issues optimistically lawsuit provide larger benefit \\ \hline
Cleaning (BERT) & if the lawsuit can be another means of focusing attention on these fundamental issues, then optimistically the lawsuit can provide a larger benefit. \\ \hline
\end{tabular}}
\end{center}
\caption{\label{font-table} Task 1 - Text cleaning examples }
\end{table}
\FloatBarrier

\section{Task 2 systems configuration.}
\subsubsection{Features.}
We apply the same feature sets to our two systems. 
The \textbf{bold} mentions refer to their description in the results Table 3 and Table 4. \\
For the BERT-MLP system, the features are introduced as additional input layers and calculated in the embedding layer. For the BiLSTM-CRF system, these features are declared in the CRF layer. \\
For \textbf{POS} tags, we use the Stanford\footnote{https://nlp.stanford.edu/software/tagger.html} Part-Of-Speech Tagger to determine the role of each word in the discourse. \\
For chunking the target sequences (\textbf{CHUNKS}), we tokenize each sequences and label each token of the chunk with its segment label (i.e., \textit{A} for Antecedent and \textit{C} for Consequent).\\
In order to generate BERT vector features (\textbf{BERTvec}), we use the BERT-as-service\footnote{https://github.com/hanxiao/bert-as-service} library (version 1.10.0) as a sentence-encoder to map variable-length sentences to fixed-length feature embeddings. \\
We also experiment with Syntactic Grammars \textbf{(SG)} by using the Stanford\footnote{https://nlp.stanford.edu/software/lex-parser.html} Parser to generate syntactic dependency relations between words. \\
Since our task requires labeling tokens, we could use a feature that references structure. As the results from \cite{Rei:17} show, the BIO tagging scheme performs consistently well for this type of task, which makes it the ideal candidate to add robustness to our feature set. Using the Stanford\footnote{https://nlp.stanford.edu/software/CRF-NER.html} Named Entity Recognizer, which is trained on BIO entity tagged data, we generate the BIO representation of sentence entities, or BIO Named Entity Recogition \textbf{(BIO NER)} tags.  

\subsubsection{Embeddings.}
\label{embed}
Embeddings are computed from convolution of the different D.0.1 features (plus additional embeddings for BiLSTM-CRF) stacked in the input layer.
The \textbf{bold} mentions refer to their description in the results Table 3 and Table 4. \\
For the \textbf{BERT-MLP} system, we add the cascaded layers of token-features referenced in section D.0.1 to the existing embeddings architecture . Our final embeddings layer is a concatenation of the 3 BERT embeddings (positional, segment, token) and the generated feature layers. \\
For feeding the BiLSTM-CRF system, we evaluate several embedding methods . \\
The first is Pre-trained Word Embeddings. This widely used technique helps tackle the problem of generalising unseen words, since word embeddings are good at capturing general syntactic as well as semantic properties of words \cite{Rei:17}. For our experiments, we focus on two different approaches: the \textbf{GloVe} embeddings trained on Common Crawl (about 840 billion tokens) and the \textbf{FastText} approach trained on Common Crawl (600 billion tokens) which also extracts subword information. \\
As per the recommended approach of \cite{Ma:16}, we also compound Character Embeddings \textbf{(C2IDX)} to take into account character-level representation of words as an additional layer to the word embedding layer. \\
Finally, we also consider Stacked Embeddings \textbf{(Stacked)} (i.e., a combination of existing embedding techniques designed to act in succession in a single pipeline to generate a refined embedding for our input). We use the Flair\footnote{https://github.com/flairNLP/flair} library (version 0.4.5) to achieve this. Our embeddings pipeline is composed of: a GloVe model, a Flair-forward model, a Flair-backward model and a BERT embedding model. By targeting these models, we cover different aspects of semantic representation for our input: the GloVe module targets word representation, the Flair modules are for character contextualisation, and the BERT layer is for sentence-level information extraction.

\subsubsection{Regularisation.}
Given the danger of over-parameterisation that neural networks present, we introduce some regularisation techniques. \\
For the BiLSTM-CRF model, we perform a \textbf{K-Fold Cross Validation} with K = 3. The data is not shuffled before splitting into batches. We also add \textbf{Dropout}. Results from \cite{Rei:17} show that variational dropout performs best when it comes to BiLSTM networks. Furthermore, it can be shown in \cite{Che:17} that relatively smaller dropout tends to yield better results for LSTM networks. For our experiments, we implement a variational dropout on all layers with the fraction \textit{p} of dropped values from the set \{0.1, 0.3, 0.5\}. The value of \textit{p} = 0.1 performs best after empirical testing and is retained for our final round of benchmark tests. We couple our system with an \textbf{Elasticnet} method (i.e., a linear regression model with combined L1 and L2 priors). 
The tested combinations of regularisation methods are the following (displayed as A, B, C in result Table 3) : 
\begin{itemize}
    \item \textbf{Cross Validation} (A)
    \item \textbf{Cross Validation + Dropout} (B)
    \item \textbf{Cross Validation + Dropout + Elasticnet} (C)
\end{itemize}

The reference labels A, B, C are used in Table 3.\\

For the BERT-MLP system, we apply the default regularisation A.

\subsubsection{Hyperparameters.}
We also evaluate the effects of different hyperparameters on the performance of our models. For the BiLSTM-CRF system, the tested configurations in Table 8 are limited to the hardware we use. The hardware specifications are: i7 CPU processor, 16 GB of RAM with a GPU of 8 GB.  
\par \textbf{Character Embedding Dimension:} We select the CNN approach recommended for deep LSTM networks \cite{Rei:17} and test configurations on the embedding size. Larger embedding dimensions cause redundancy while capturing character-level information and perform worse than smaller sizes.
\par \textbf{Word Embedding Dimension:} For word-level embeddings, we also test different configurations and their impact on the model. Since we also make use of pre-trained embedding models, we observe that setting the word dimension to a size close to the usual pre-trained word vectors (100 - 300) yields the best performing models.
\par \textbf{BiLSTM Layer Dimension:} We evaluate 1, 2 and 3 stacked BiLSTM layers.
\par \textbf{Recurrent Units:} The number of recurrent units \textit{u} is selected from a range of sizes with 32 $\leq$ u $\leq$ 100. The forward and reverse running LSTM networks have the same number of recurrent units. Multiple BiLSTM layers also have the same number of recurrent units.
\par \textbf{Optimizer:} We experiment with two commonly selected optimizers, namely stochastic gradient descent (\textbf{SGD}) and \textbf{Adamax}. SGD is highly sensitive to the learning rate, meaning choosing a too high rate can cause the system to diverge in terms of the objective function, whereas a too low rate results in a slow learning process. Adamax is selected to bypass the shortcomings of SGD.
\par \textbf{Learning Rate:} We tune the learning rate by hand and observe in many instances that it fails to converge to a minimum. We select a range of viable rates to test for best performance.
\par \textbf{Weight Decay (L2 Penalty):} We couple weight decay with our optimizer to add regularisation and evaluate the best value for our model.

\begin{table}[h]
\begin{center}
\begin{tabular}{|l|r|r|r|r|r|}
\hline \bf Parameter & \bf Tested Configurations & \bf Best Configuration \\ \hline
Character Embedding Dimension & {25, 50, 75, 100} & 25 \\
Word Embedding Dimension & {50, 200, 300, 500} & 300 \\
BiLSTM Layer Dimension & {1, 2, 3} & 3 \\
Recurrent Units & {32, 50, 64, 75, 100} & 100 \\
Optimizer & {Adamax, SGD} & Adamax \\
Learning Rate & {0.0015, 0.002, 0.015, 0.02} & 0.0015 \\
Weight Decay (L2 Penalty) & {$15\mathrm{e}{-10}$, $15\mathrm{e}{-12}$, $15\mathrm{e}{-14}$} & $15\mathrm{e}{-12}$ \\
\hline
\end{tabular}
\end{center}
\caption{\label{font-table} Task 2 - Hyperparameter Evaluation}
\end{table}

The best configuration for each parameter in Table 8 is considered the optimal one for our use case and our model is tuned to those values for the benchmark tests.

For parameter tuning in the BERT-MLP case, we use a Random Search to facilitate multi-parameter testing. The optimal parameter selection is represented below:
\par \textbf{Number of MLP Layers:} 3
\par \textbf{Number of MLP Hidden Neurons:} 500
\par \textbf{Max Sequence Length:} 173
\par \textbf{Number of Epochs:} 3
\par \textbf{Learning Rate:} $5\mathrm{e}{-5}$
\par \textbf{Batch Size:} 16
\par \textbf{Gradient Accumulation Steps:} 2

\section{Architecture of Task 2 BiLSTM system.}
\FloatBarrier
\begin{figure}
    \centering
    \includegraphics[width=\textwidth]{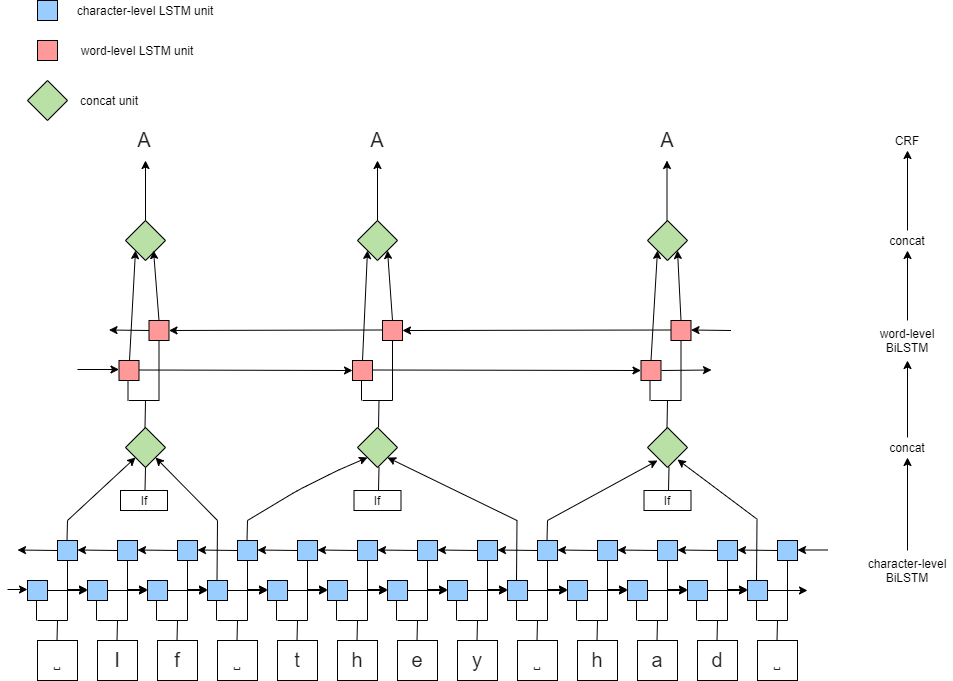}
    \caption{Task 2 - BiLSTM-CRF architecture}
\end{figure}
\FloatBarrier
\end{appendices}

\end{document}